\documentclass[10pt,twocolumn]{article}

\usepackage[a4paper,margin=0.7in]{geometry}
\usepackage[T1]{fontenc}
\usepackage{lmodern}

\usepackage{graphicx}
\usepackage{booktabs}
\usepackage{multirow}
\usepackage{dblfloatfix}
\usepackage{amsmath}
\usepackage{amssymb}
\usepackage{mathtools}
\usepackage{microtype}
\usepackage{xcolor}
\usepackage{cite}
\usepackage[hidelinks]{hyperref}
\usepackage{titling}

\pretitle{\centering\Large\bfseries}
\posttitle{\par\vskip 0.5em}

\title{
\textbf{PerSeM: Persistent Semantic Memory for Long-Horizon\\
Open-Vocabulary UAV Mapping}
}

\author{
Saurbh Singh Jamwal, Ganesh Ramakrishnan\\
Department of Computer Science and Engineering\\
Indian Institute of Technology Bombay\\
\texttt{\{saurbh,ganesh\}@cse.iitb.ac.in}
}

\date{}

\begin{document}
\raggedbottom

\maketitle

\begin{abstract}

Open-vocabulary segmentation enables rich semantic perception for UAVs, but frame-wise predictions can remain temporally inconsistent across repeated observations and changing viewpoints. We present PerSeM, a training-free persistent semantic memory framework for long-horizon open-vocabulary UAV mapping. PerSeM associates frame-wise semantic observations with persistent world-space voxels and constructs a majority-based semantic memory, which is conservatively refined through history-preserving spatial refinement, trust-aware replay, and context-guided verification. Experiments on the Forest and UAVScenes benchmarks show that persistent 3D memory provides substantial gains in semantic correctness and temporal stability over frame-wise predictions. Beyond this strong persistent-memory baseline, PerSeM provides consistent additional improvements, improving both semantic accuracy and temporal stability across all five evaluated UAVScenes sequences. Analysis using regions identified independently of the final PerSeM predictions further shows that these gains are concentrated in semantically difficult and temporally unstable regions, where majority-based memory is most likely to remain uncertain. These results demonstrate that persistent 3D aggregation provides a strong foundation for long-horizon semantic mapping, while conservative refinement of uncertain memory states can provide additional improvements without retraining or additional neural-network inference.

\end{abstract}

\vspace{0.5em}

\section{Introduction}
\label{sec:introduction}

Recent advances in open-vocabulary semantic segmentation and large-scale vision foundation models have significantly improved semantic understanding in UAV imagery, enabling downstream robotic capabilities such as autonomous navigation, active perception, traversability reasoning, and obstacle-aware planning. By leveraging large-scale visual and language priors, modern open-vocabulary models can recognise diverse semantic concepts without task-specific retraining, making them particularly attractive for long-horizon UAV perception in complex environments.

Despite these advances, semantic predictions in UAV video streams often remain temporally unstable under viewpoint variation, sparse revisitation, and visually ambiguous scene content. Our analysis reveals that semantic fluctuations are concentrated within a relatively small subset of recurrent scene regions, where frame-wise predictions fail to maintain persistent semantics in world space. Existing approaches for temporal consistency primarily operate in the 2D image or video domain through temporal smoothing, feature correspondence, and short-term memory mechanisms. While these methods improve visual coherence between neighboring frames, robotic interaction occurs within persistent 3D environments, where semantic information must be accumulated and maintained across multiple observations and viewpoint changes.

A key observation of this work is that persistent world-space aggregation resolves much of this frame-level instability: repeated observations establish reliable semantic assignments for most recurrent regions. The remaining challenge is not global semantic aggregation, but identifying and correcting uncertain memory states without disturbing stable assignments. We introduce \textbf{PerSeM}, a persistent semantic memory framework designed as a conservative refinement layer over persistent 3D memory. PerSeM associates semantic observations with world-space voxels, establishes a memory from repeated observations, and selectively revisits uncertain states using history-preserving spatial refinement, trust-aware evidence replay, and contextual verification. This preserves stable assignments while allowing supported evidence to correct residual errors. Experiments on Forest and UAVScenes confirm that persistent 3D memory provides the dominant improvement over frame-wise predictions, while PerSeM further improves semantic correctness and temporal stability beyond majority memory and alternative persistent fusion strategies. Our main contributions are summarised as:

\begin{itemize}

\item We introduce \textbf{PerSeM}, a persistent semantic memory framework for open-vocabulary UAV mapping, and identify correction of residual uncertain memory states as a key challenge after persistent world-space aggregation.

\item We propose a conservative memory refinement strategy that preserves accumulated semantic history and combines context-guided verification with trust-aware residual proposals, allowing uncertain assignments to be revised while preserving stable memory states.

\item We evaluate PerSeM on Forest and five UAVScenes sequences against frame-wise, image-space temporal, majority-memory, and alternative persistent fusion baselines. We further analyse independently identified difficult regions, component roles, parameter sensitivity, and failure modes to characterise when refinement provides benefits beyond persistent aggregation.

\end{itemize}
\section{Related Work}
\label{sec:background}

\paragraph{Open-Vocabulary Semantic Mapping}

Recent advances in foundation models have significantly expanded open-vocabulary semantic perception across natural and remote sensing imagery. Segment Anything (SAM)~\cite{kirillov2023segment} enables large-scale promptable segmentation, while SegEarth-OV~\cite{li2025segearthov} extends foundation-model capabilities to training-free open-vocabulary semantic segmentation for aerial and remote sensing imagery using vision-language priors.

Building upon these semantic observations, several works have explored open-vocabulary spatial representations. OpenScene~\cite{peng2023openscene} introduced language-aligned 3D scene representations through spatially consistent semantic features, while ConceptFusion~\cite{jatavallabhula2023conceptfusion} demonstrated persistent multimodal 3D mapping by integrating foundation-model semantics into scene representations. These works demonstrate the potential of combining open-vocabulary semantics with persistent spatial representations. In contrast, PerSeM treats open-vocabulary segmentation outputs as semantic observations and focuses on maintaining temporally consistent semantic memory for long-horizon mapping.

\paragraph{Temporal Consistency in Video Perception}

Temporal consistency has been studied in video segmentation, where the objective is to maintain semantic predictions across sequences. MaskProp~\cite{bertasius2020maskprop}, STCN~\cite{cheng2021rethinking}, AOT~\cite{yang2021associating}, XMem~\cite{cheng2022xmem}, and Video K-Net~\cite{li2022videoknet} improve temporal coherence through mask propagation, feature correspondence, transformer-based association, and memory mechanisms.


While these approaches effectively reduce temporal instability in image sequences, they operate primarily in image space and model short- to medium-term temporal relationships. In contrast, PerSeM formulates temporal consistency as a persistent world-space problem, where semantic observations are accumulated, revisited, and selectively refined within a long-term voxel memory to improve semantic stability over extended UAV trajectories.

\paragraph{Persistent Semantic Mapping and Spatial Memory}

Most existing semantic mapping systems emphasize semantic completeness, representation quality, or queryable open-vocabulary scene representations. In contrast, PerSeM studies the temporal stability of persistent semantic memories over long-horizon UAV observations.

Persistent spatial representations have become important for long-horizon scene understanding and robotic perception. SemanticFusion~\cite{mccormac2017semanticfusion} pioneered dense semantic mapping by integrating semantic predictions into persistent volumetric maps. Subsequent systems including PanopticFusion~\cite{narita2019panopticfusion}, Kimera~\cite{rosinol2021kimera}, and Hydra~\cite{hughes2022hydra} demonstrated the value of persistent semantic world models for long-term perception, scene understanding, and robot interaction.

More recently, vision-language models have enabled open-vocabulary semantic mapping beyond fixed category sets. VLMaps~\cite{huang2023vlmaps}, OpenScene~\cite{peng2023openscene}, Open-Fusion~\cite{yamazaki2024openfusion}, and ConceptFusion~\cite{jatavallabhula2023conceptfusion} introduced language-grounded spatial representations through dense semantic features, TSDF fusion, and multimodal scene representations. Recent systems have further improved open-vocabulary mapping through probabilistic voxel representations, instance-level reasoning, and hybrid semantic memories. OpenVox~\cite{deng2025openvox} performs uncertainty-aware probabilistic voxel fusion for real-time open-vocabulary mapping, OVI-MAP~\cite{deng2026ovimap} constructs persistent instance-semantic maps through multi-view instance association, and FUS3DMaps~\cite{homberger2026fus3dmaps} combines dense voxel and instance-level semantic embeddings using dual-layer fusion for scalable open-vocabulary mapping. OVO~\cite{martins2025openvocabulary} further integrates persistent object tracking, CLIP-based semantic descriptors, and SLAM for online open-vocabulary semantic mapping.

Unlike these approaches, which focus on improving semantic representations through probabilistic fusion, instance reasoning, embedding aggregation, or semantic retrieval, PerSeM targets temporal semantic stability as a first-class objective. Rather than redesigning the semantic representation, PerSeM improves the temporal reliability of persistent voxel memories through history-preserving aggregation, trust-aware replay, and context-guided verification.

\paragraph{Semantic Perception for UAV Autonomy}

Semantic scene understanding has become important for UAV navigation, active perception, obstacle avoidance, and autonomous decision making. Prior work has incorporated semantic information into perception-aware path planning~\cite{bartolomei2020perceptionaware}, semantic-aware viewpoint selection~\cite{bartolomei2021semanticaware}, obstacle-aware trajectory optimisation~\cite{xu2022visionaided}, and semantic-driven UAV navigation~\cite{yue2024semanticnav}. These studies demonstrate the role of semantic perception in autonomous UAV operation.

\begin{figure*}[t]
\centering
\includegraphics[width=\textwidth]{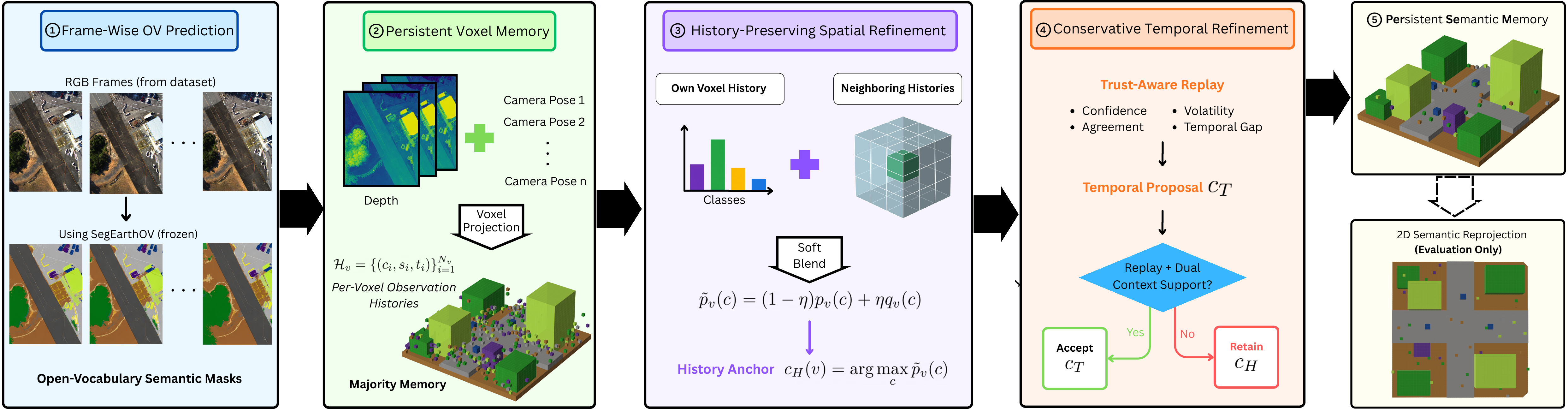}
\caption{
Overview of the PerSeM framework. Frame-wise semantic observations are associated with persistent world-space voxels and accumulated into observation histories. Starting from majority-based voxel memory, PerSeM constructs a history-preserving spatial anchor and selectively applies trust-aware temporal proposals that satisfy conservative replay and context-guided verification criteria.
}
\label{fig:architecture}
\end{figure*}

However, most existing UAV systems rely on frame-wise semantic predictions that are susceptible to viewpoint changes, occlusions, and sparse scene revisitation. PerSeM addresses this limitation by maintaining a persistent voxel-level semantic memory that reinforces reliable semantic observations across repeated views and long revisit intervals, improving long-horizon semantic consistency while preserving semantic correctness.

\section{Methodology}
\label{sec:method}

\subsection{Overview}

We propose PerSeM, a persistent semantic memory framework for temporally consistent open-vocabulary UAV mapping. Given frame-wise semantic predictions, depth estimates, and camera poses, PerSeM associates semantic observations with persistent world-space voxel locations rather than processing images independently. Each voxel maintains an observation history across repeated views, enabling semantic evidence to accumulate over time while preserving the temporal structure of observations.

Figure~\ref{fig:architecture} illustrates the pipeline. PerSeM first constructs a majority-based persistent voxel memory, which provides a semantic anchor for stable locations. A history-preserving spatial refinement then combines each voxel's accumulated semantic distribution with those of its neighbors to obtain locally supported corrections without discarding its observation history. Trust-aware evidence replay subsequently generates temporal correction proposals for recurrent voxels, which are accepted only when they provide sufficient evidence over the majority assignment and are supported by the surrounding majority memory and neighboring semantic-history distributions. This conservative design preserves stable memory assignments while selectively revisiting uncertain semantic states. Dense image-space reprojection is used only for evaluation and visualisation.

\subsection{Persistent Voxel Memory}

Given frame-wise semantic predictions, depth estimates, and camera poses, PerSeM backprojects valid semantic observations into world coordinates and associates them with a voxelised 3D representation. Observations from different frames and viewpoints are therefore associated whenever they correspond to the same persistent world-space voxel. To prevent densely sampled image regions from dominating the memory, observations assigned to the same voxel within a frame are aggregated into a single semantic observation using majority voting, with confidence given by the corresponding within-voxel semantic agreement. 

Each voxel $v$ consequently maintains a temporally ordered observation history

\begin{equation*}
\mathcal{H}_v
=
\{(c_i,s_i,t_i)\}_{i=1}^{N_v},
\end{equation*}

where $c_i$, $s_i$, and $t_i$ denote the semantic label, confidence, and frame index of observation $i$, respectively, and $N_v$ is the number of times the voxel has been observed. The accumulated class evidence is represented by

\begin{equation*}
h_v(c)
=
\sum_{i=1}^{N_v}
s_i\,\mathbf{1}[c_i=c],
\end{equation*}

and the initial persistent-memory label is obtained as

\begin{equation*}
c_M(v)
=
\arg\max_c h_v(c).
\end{equation*}

This majority-based memory provides the semantic anchor from which PerSeM performs subsequent refinement. Voxels with insufficient observation history retain this assignment, while recurrent voxels expose both the accumulated class evidence and ordered observation history to the history-preserving spatial and trust-aware refinement stages described below.

\subsection{History-Preserving Spatial Refinement}

Majority memory treats each voxel independently and may retain locally inconsistent assignments when its observation history is ambiguous. PerSeM therefore incorporates neighboring semantic histories while preserving each voxel's accumulated evidence. We first normalise the accumulated class evidence into probabilities as

\begin{equation*}
p_v(c)
=
\frac{h_v(c)}
{\sum_{c'} h_v(c')},
\end{equation*}

and compute the mean history distribution $q_v(c)$ over the spatial neighborhood $\mathcal{N}(v)$. The two distributions are softly combined as

\begin{equation*}
\begin{aligned}
\tilde{p}_v(c)
&=
(1-\eta)p_v(c)+\eta q_v(c),\\
q_v(c)
&=
\frac{1}{|\mathcal{N}(v)|}
\sum_{u\in\mathcal{N}(v)} p_u(c).
\end{aligned}
\end{equation*}

where $\eta$ controls the contribution of neighboring histories. The history-preserving anchor is then $c_H(v)=\arg\max_c\tilde{p}_v(c)$.

Unlike hard neighborhood voting, this refinement introduces local support without discarding the voxel's accumulated semantic evidence. Voxels modified at this stage are protected from subsequent trust-based overwriting, such that trust replay acts only as a conservative residual correction to the resulting memory.

\subsection{Trust-Aware Evidence Replay}

For recurrent voxels whose history anchor retains the majority-memory assignment, PerSeM uses the ordered observation history to determine whether temporal evidence supports an alternative label. Observations are replayed sequentially, with each observation $o_i$ assigned a trust weight based on its confidence, class volatility, semantic agreement, and temporal gap:

\begin{equation*}
w_i
=
s_i
\exp(-\lambda_v \rho(c_i))
\exp(-\lambda_s g_i)
\exp(-\lambda_t \Delta t_i),
\end{equation*}

where $s_i$ is the observation confidence, $\rho(c_i)$ denotes class volatility, $g_i$ indicates disagreement with the current dominant belief, and $\Delta t_i$ is the temporal gap from the preceding observation.

During replay, the accumulated class evidence is exponentially decayed and updated as

\begin{equation*}
\mathbf{e}_i=\gamma\mathbf{e}_{i-1},
\qquad
e_i(c_i)\leftarrow e_i(c_i)+w_i,
\end{equation*}

where $\gamma$ controls evidence decay. After processing the complete history, replay produces the temporal proposal

\begin{equation*}
c_T(v)=\arg\max_c e(c).
\end{equation*}

Unlike unrestricted trust-based relabeling, $c_T(v)$ is treated only as a correction proposal. A proposal that disagrees with the majority-memory label is considered further only when its replay evidence exceeds that of the majority label by a margin $\tau_{\mathrm{replay}}$; otherwise, the history anchor is retained. Surviving proposals are subsequently subjected to context-guided verification.

\subsection{Conservative Context Verification}

Trust proposals that pass the replay-margin criterion are accepted only when they are also supported by the surrounding semantic memory. PerSeM evaluates two complementary forms of spatial evidence: the hard majority labels of neighboring voxels and their accumulated semantic-history distributions.

For a candidate label $c$, hard contextual support from neighboring majority assignments is computed as

\begin{equation*}
S_{\mathrm{hard}}(v,c)
=
\frac{1}{|\mathcal{N}(v)|}
\sum_{u\in\mathcal{N}(v)}
\mathbf{1}[c_M(u)=c],
\end{equation*}

while history-based support is obtained from the neighboring distributions as

\begin{equation*}
S_{\mathrm{hist}}(v,c)
=
\frac{1}{|\mathcal{N}(v)|}
\sum_{u\in\mathcal{N}(v)}p_u(c).
\end{equation*}

A trust proposal $c_T(v)$ replaces the majority assignment $c_M(v)$ only when both contextual margins provide sufficient support for the proposal,

\begin{equation*}
\begin{aligned}
S_{\mathrm{hard}}(v,c_T)-S_{\mathrm{hard}}(v,c_M)
&> \tau_{\mathrm{context}},\\
S_{\mathrm{hist}}(v,c_T)-S_{\mathrm{hist}}(v,c_M)
&> \tau_{\mathrm{context}}.
\end{aligned}
\end{equation*}

Otherwise, the history-preserving anchor $c_H(v)$ is retained. Together with the replay-margin gate and protection of voxels already modified by spatial refinement, this dual verification restricts trust-based corrections to locations where temporal and spatial evidence agree, yielding a conservative refinement of the persistent memory.

\subsection{Dense Semantic Reprojection}

PerSeM operates entirely on the persistent world-space voxel memory. For evaluation and visualisation, the final voxel labels are reprojected into each camera view using the corresponding depth observations, camera intrinsics, and poses. The same voxel quantisation used during memory construction consistently associates image observations with the refined persistent memory labels, producing dense semantic maps for comparison with frame-level ground truth. Reprojection is used only for semantic evaluation, temporal stability analysis, and qualitative visualisation, and does not modify the persistent memory.

\section{Experiments}
\label{sec:experiments}

\subsection{Experimental Setup}

We evaluate PerSeM on the \textbf{Forest}~\cite{blaga2024forestinspection} and \textbf{UAVScenes}~\cite{wang2025uavscenes} semantic video benchmarks. Forest evaluates long-term semantic stability in outdoor aerial scenes, while UAVScenes provides frame-level semantic ground truth for quantitative evaluation. Following prior open-vocabulary mapping practice, we additionally report grouped-taxonomy results on UAVScenes to reduce dataset--model taxonomy mismatch.

We compare PerSeM against frame-wise \textbf{Raw2D}, image-space exponential moving average smoothing (\textbf{EMA2D}), and five persistent 3D fusion strategies operating on the same observations and voxel representation. \textbf{Raw3D Majority} assigns each voxel its most frequently observed label, while \textbf{Unweighted Majority} performs equivalent hard voting without confidence weighting. \textbf{Recency-Weighted} gives greater weight to recent observations, \textbf{Consensus + Recency} additionally favors observations agreeing with the current consensus, and \textbf{Pseudo-Probabilistic} accumulates confidence-weighted class evidence. PerSeM instead selectively refines persistent memory using history-preserving spatial refinement followed by trust-aware evidence replay and context verification; its individual components are evaluated separately in the ablation study. The EMA decay parameter was selected using a small validation sweep, with details reported in the supplementary material. Performance is evaluated using \textbf{Pixel Accuracy (PA)}, \textbf{mIoU}, and \textbf{Flicker}. PA and mIoU measure semantic correctness, while Flicker measures the percentage of valid pixels whose semantic label changes between consecutive frames, with lower values indicating greater temporal consistency. All methods use identical semantic observations, geometry, voxelisation, and benchmark splits.

\subsection{Main Results}

Table~\ref{tab:main_results} compares semantic correctness and temporal stability on Forest and UAVScenes. We include frame-wise predictions (Raw2D), image-space temporal smoothing (EMA2D), persistent Raw3D Majority, and alternative semantic-fusion strategies operating on the same observations and voxel representation. Results are macro-averaged across two Forest sequences and five UAVScenes sequences using a single frozen configuration.

\begin{table*}[t]
\centering
\caption{Semantic correctness and temporal stability on Forest and UAVScenes. All persistent-memory methods use identical frame-wise semantic observations, geometry, and voxelisation. UAVScenes results are macro-averaged across five sequences.}
\label{tab:main_results}
\setlength{\tabcolsep}{5pt}
\begin{tabular}{lcccccc}
\toprule
& \multicolumn{3}{c}{Forest} & \multicolumn{3}{c}{UAVScenes} \\
\cmidrule(lr){2-4} \cmidrule(lr){5-7}
Method &
PA $\uparrow$ & mIoU $\uparrow$ & Flicker $\downarrow$ &
PA $\uparrow$ & mIoU $\uparrow$ & Flicker $\downarrow$ \\
\midrule
Raw2D
& 75.00 & 37.31 & 37.49
& 77.50 & 36.32 & 13.73 \\

EMA2D
& 80.43 & 24.45 & \textbf{10.21}
& 78.56 & 36.29 & \textbf{6.86} \\
\midrule
Raw3D Majority
& 83.44 & 39.66 & 24.39
& 81.21 & 41.50 & 7.27 \\

Unweighted Majority
& 83.31 & 39.81 & 24.98
& 81.21 & 41.48 & 7.27 \\

Recency-Weighted
& 80.85 & 39.64 & 29.53
& 80.07 & 38.56 & 8.22 \\

Consensus + Recency
& 81.15 & \textbf{40.08} & 29.05
& 80.03 & 38.50 & 8.22 \\

Pseudo-Probabilistic
& 83.64 & 39.61 & 23.73
& 81.22 & 41.58 & 7.25 \\
\midrule
PerSeM
& \textbf{84.02} & 39.84 & 23.00
& \textbf{81.41} & \textbf{41.90} & 6.90 \\
\bottomrule
\end{tabular}
\end{table*}

Across both benchmarks, persistent 3D aggregation provides the largest improvement over frame-wise predictions, indicating that repeated world-space observations already resolve much of the semantic instability. The remaining headroom is therefore concentrated in residual uncertain memory states rather than the full persistent map. EMA2D strongly suppresses flicker but can sacrifice semantic accuracy, while alternative 3D fusion strategies remain near Raw3D Majority or improve one metric at the expense of others. PerSeM instead preserves stable majority-memory assignments and selectively revisits uncertain states using historical reliability and context-guided support.

This selective refinement consistently improves upon Raw3D Majority. On Forest, PerSeM improves PA from 83.44 to 84.02 and mIoU from 39.66 to 39.84 while reducing flicker from 24.39 to 23.00. Although Consensus + Recency reaches 40.08 mIoU, it incurs substantially lower PA and higher flicker. On UAVScenes, PerSeM improves PA from 81.21 to 81.41 and mIoU from 41.50 to 41.90 while reducing flicker from 7.27 to 6.90. PerSeM is the only evaluated persistent-memory variant that improves all three metrics over Raw3D Majority at the macro level on both benchmarks, with mIoU and flicker improvements holding across all five UAVScenes sequences. Section~\ref{sec:why_persem_helps} shows that the improvement is further concentrated in independently identified ambiguous and temporally unstable regions, consistent with PerSeM's intended role as a selective refinement mechanism for residual memory uncertainty.

\subsection{Stability--Correctness Tradeoff}

\begin{figure}[t]
\centering
\includegraphics[width=0.97\linewidth]{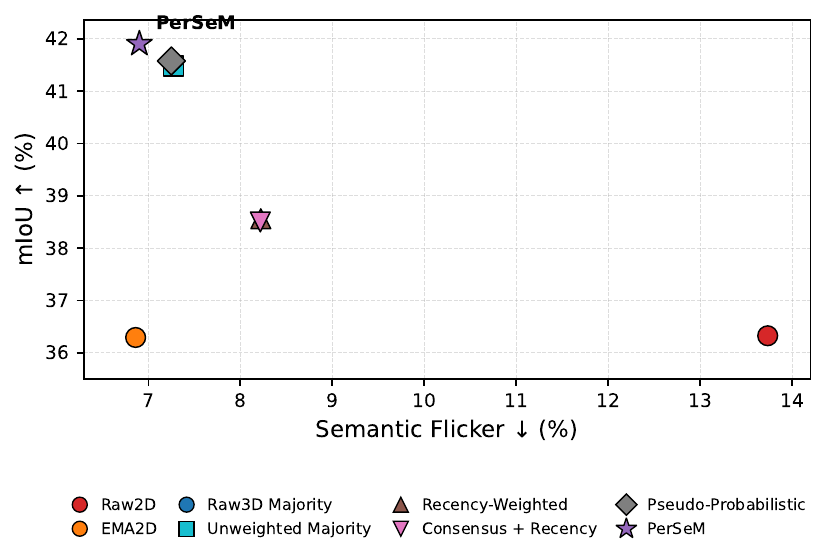}
\caption{
Semantic stability--correctness tradeoff on the UAVScenes benchmark. PerSeM achieves the best balance between semantic correctness and temporal stability.
}
\label{fig:uavscenes_tradeoff}
\end{figure}

Figure~\ref{fig:uavscenes_tradeoff} visualises the relationship between semantic correctness and temporal stability on UAVScenes. Raw2D exhibits substantial temporal instability, while EMA2D strongly reduces flicker but provides limited semantic improvement. Persistent 3D fusion shifts the operating point toward higher semantic correctness and greater stability, demonstrating the benefit of associating repeated observations in world space.

Among persistent-memory methods, alternative voting, recency, consensus, and confidence-weighted fusion strategies remain within a relatively narrow performance region. PerSeM moves beyond this region toward higher mIoU and lower flicker by selectively refining residual uncertain memory states. This separates the primary gain from persistent 3D aggregation from the additional improvement obtained through selective memory refinement, without relying on aggressive temporal smoothing that can reduce flicker at the expense of semantic correctness.

\subsection{Component Ablation}

To isolate the roles of the refinement components, we evaluate Raw3D Majority, Trust Only, Context Only, History Blend Only, and the full PerSeM pipeline on the UAVScenes grouped benchmark. Trust Only applies reliability-aware temporal evidence replay without contextual verification, Context Only performs context-guided refinement without trust replay, and History Blend Only retains accumulated observation history as a soft spatial anchor. PerSeM Full combines history-preserving refinement, trust-aware replay, and contextual verification. Results are summarised in Table~\ref{tab:uavscenes_ablation}.

\begin{table}[t]
\centering
\caption{Component ablation on the UAVScenes grouped benchmark across five evaluated sequences.}
\label{tab:uavscenes_ablation}
\begin{tabular}{@{}lccc@{}}
\toprule
Variant & PA $\uparrow$ & mIoU $\uparrow$ & Flicker $\downarrow$ \\
\midrule
Raw3D Majority     & 81.21 & 41.50 & 7.27 \\
Trust Only         & 80.31 & 38.36 & 8.27 \\
Context Only       & \textbf{81.47} & 41.88 & \textbf{6.41} \\
History Blend Only & 81.26 & 41.59 & 7.11 \\
PerSeM Full        & 81.41 & \textbf{41.90} & 6.90 \\
\bottomrule
\end{tabular}
\end{table}

Raw3D Majority provides a strong baseline by aggregating repeated observations into persistent voxel assignments. Trust replay in isolation degrades all three metrics, showing that temporal reliability alone is insufficient to replace an established majority state. In contrast, Context Only improves all three metrics, while History Blend Only provides smaller but consistent gains. These results identify persistent history and contextual support as reliable refinement cues and motivate constraining trust replay rather than allowing it to determine the persistent label.

The aggregate ablation, however, does not capture the scene-dependent role of trust replay. We therefore compare Context+History against the complete pipeline on two representative UAVScenes sequences in Table~\ref{tab:trust_tradeoff} to illustrate this semantic accuracy--stability trade-off more directly. These cases expose when residual temporal evidence helps or harms an established persistent state. Adding trust replay improves semantic accuracy on HKisland01, but perturbs an already stable solution on HKairport01.

\begin{table}[t]
\centering
\caption{Scene-dependent effect of adding trust replay to Context+History across representative UAVScenes sequences.}
\label{tab:trust_tradeoff}
\setlength{\tabcolsep}{4pt}
\resizebox{\columnwidth}{!}{%
\begin{tabular}{@{}llccc@{}}
\toprule
Scene & Variant & PA $\uparrow$ & mIoU $\uparrow$ & Flicker $\downarrow$ \\
\midrule
HKisland01
 & Context+History & 58.55 & 25.28 & \textbf{5.15} \\
 & Full (+Trust) & \textbf{59.05} & \textbf{26.14} & 6.12 \\
\midrule
HKairport01
 & Context+History & \textbf{88.16} & \textbf{48.46} & \textbf{5.78} \\
 & Full (+Trust) & 87.72 & 48.18 & 6.37 \\
\bottomrule
\end{tabular}%
}
\end{table}

On HKisland01, trust replay recovers $+0.50$ PA and $+0.86$ mIoU relative to Context+History, although flicker increases by $0.97$ points. Conversely, on HKairport01, adding trust reduces PA and mIoU by $0.44$ and $0.28$ points and increases flicker by $0.59$ points. Trust replay should therefore not be interpreted as a uniformly beneficial estimator. Instead, it provides a residual recovery pathway that can recover semantic evidence when the persistent state remains unreliable, while contextual verification limits its influence when that evidence is unsupported by the surrounding persistent memory. This scene-dependent trade-off motivates retaining trust as a constrained component of the complete pipeline rather than attributing PerSeM's overall improvement to trust weighting alone.

\subsection{Where Does PerSeM Help?}
\label{sec:why_persem_helps}

The gains over Raw3D Majority are modest when averaged over the full scene because persistent majority fusion already provides stable assignments in many regions. To test whether PerSeM specifically benefits difficult memory states, we define challenging regions independently of the final PerSeM output. We consider: (i) \textbf{boundary regions}, where the local Raw3D neighborhood contains multiple semantic labels; (ii) \textbf{high-switch regions}, where frame-wise semantic observations frequently change labels over time; and (iii) a \textbf{difficult core}, where at least two of four pre-refinement ambiguity cues---high observation entropy, low majority margin, high switch rate, and proximity to a semantic boundary---are present. These regions are identified using only observation histories and Raw3D spatial structure; neither ground truth nor PerSeM predictions are used for their selection. Exact definitions and thresholds are provided in the supplementary material.

\begin{table}[t]
\centering
\small
\setlength{\tabcolsep}{4pt}
\caption{Performance gains over Raw3D Majority in independently identified difficult regions on UAVScenes. Regions are defined before PerSeM refinement.}
\label{tab:difficult_region_analysis}
\begin{tabular}{lcccc}
\toprule
Region & Pixels (\%) & $\Delta$PA & $\Delta$mIoU & $\Delta$Flicker \\
\midrule
All valid       & 100.00 & +0.20 & +0.41 & -0.37 \\
Boundary        & 48.13  & +0.38 & +0.54 & -0.80 \\
Difficult core  & 38.88  & +0.53 & +0.52 & -0.92 \\
High switch     & 8.28   & +1.22 & +0.28 & -1.80 \\
\bottomrule
\end{tabular}
\end{table}

Table~\ref{tab:difficult_region_analysis} shows that the modest aggregate improvement masks substantially larger gains in regions that are difficult before refinement. At semantic boundaries, the PA gain increases from $+0.20$ to $+0.38$ points and the flicker reduction from $0.37$ to $0.80$ points, while the difficult core reaches $+0.53$ PA and a $0.92$-point flicker reduction. The effect is most pronounced for temporal stability in high-switch regions, which comprise only $8.28\%$ of evaluated pixels: PerSeM improves PA by $1.22$ points and reduces flicker by $1.80$ points, while retaining a positive $+0.28$ mIoU gain. These results support the motivation for selective refinement: majority fusion is already effective over much of the scene, whereas retaining and selectively reconsidering observation history provides greater benefit in ambiguous and temporally inconsistent memory states.

\subsection{Qualitative Results}

\begin{figure*}[t]
\centering
\includegraphics[width=\textwidth]{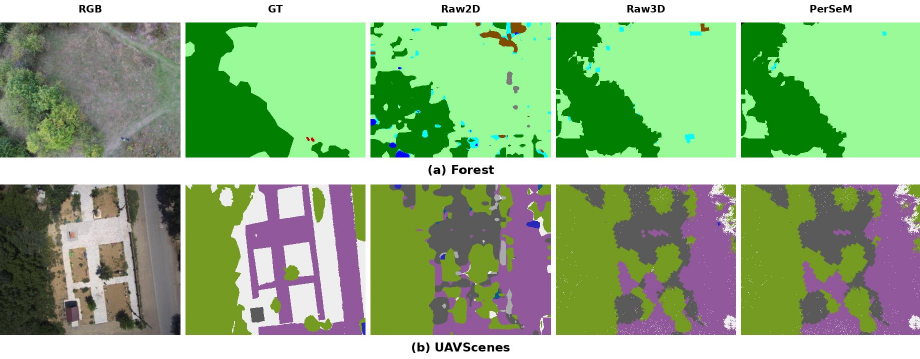}
\caption{
Qualitative semantic mapping results on (a) Forest and (b) UAVScenes. Raw3D majority memory resolves much of the fragmentation present in frame-wise Raw2D predictions. PerSeM subsequently performs localized refinements of residual ambiguous memory states while largely preserving already coherent persistent assignments.
}
\label{fig:qualitative_results}
\end{figure*}

Figure~\ref{fig:qualitative_results} presents representative examples from the Forest and UAVScenes benchmarks. Raw2D predictions exhibit substantial local fragmentation, whereas persistent majority aggregation in Raw3D produces considerably more coherent, temporally stable semantic regions. This visually reinforces our quantitative finding that persistent world-space aggregation accounts for the dominant reduction in frame-level semantic instability across repeated observations.

PerSeM operates conservatively on this persistent representation rather than broadly altering the resulting map. In both examples, the overall semantic structure established by Raw3D is retained, while the differences introduced by PerSeM are concentrated in localized regions. The Forest example shows refinement of small residual inconsistencies within otherwise coherent vegetation regions, while the UAVScenes example demonstrates similarly localized changes along heterogeneous surface and structure regions.

These examples complement the quantitative analysis in Section~\ref{sec:why_persem_helps}: persistent 3D aggregation provides the dominant first improvement, while PerSeM focuses its refinement on the smaller set of residual uncertain memory states where accumulated history and local contextual evidence can reliably support a semantic update.

\subsection{Runtime and Memory Analysis}

Table~\ref{tab:runtime_storage} reports the computational cost of PerSeM on Forest and UAVScenes. We separate shared voxel-history construction from PerSeM-specific semantic refinement, while excluding the common open-vocabulary segmentation inference. Reprojection is reported separately because it is required only to generate dense image-space predictions for quantitative evaluation and is not needed to maintain the persistent voxel memory.

\begin{table}[t]
\centering
\caption{Runtime and peak memory of PerSeM, averaged across all evaluated benchmark sequences. Reprojection is used only for generating image-space evaluation outputs.}
\label{tab:runtime_storage}
\resizebox{\columnwidth}{!}{%
\begin{tabular}{@{}lrrrrr@{}}
\toprule
Dataset & Voxels & History & Refinement & Reproj. & Peak Mem. \\
& & (s) & (s) & (s) & (GB) \\
\midrule
Forest    & 19.4K  & 114.6 & 2.82  & 1748.0 & 0.93 \\
UAVScenes & 291.8K & 54.5  & 39.10 & 353.2  & 3.17 \\
\bottomrule
\end{tabular}%
}
\end{table}

PerSeM's method-specific refinement requires 2.82\,s on Forest and 39.10\,s on UAVScenes. Compared with Raw3D Majority, the complete pipeline runtime increases by only 0.13\% (1863.1$\rightarrow$1865.4\,s) and 8.4\% (412.2$\rightarrow$446.7\,s), respectively. Reprojection dominates the measured cost, particularly on Forest, but is required only to produce image-space predictions for evaluation rather than to maintain persistent memory. Detailed comparisons with alternative fusion baselines are provided in the supplementary material.

Peak memory usage is 0.93\,GB on Forest and 3.17\,GB on UAVScenes. PerSeM operates on accumulated voxel-history statistics and requires no additional neural inference, retraining, or external semantic model beyond the shared open-vocabulary predictions. These measurements use an unoptimised research implementation and therefore characterise conservative refinement overhead rather than fully optimised deployment performance in practice.

\section{Limitations and Future Work}
\label{sec:limitations}

PerSeM is a conservative refinement of persistent majority memory and inherits some of its limitations. Stable but incorrect majority assignments may remain unchanged, while sparse or minority classes can be absorbed by dominant observations. History-aware blending provides a softer mechanism for revisiting such assignments by incorporating neighborhood evidence, but fixed blending can favor dominant local semantics and may not preserve rare or spatially small classes. This is relevant for safety-critical categories, motivating future work on adaptive history blending, class-aware preservation, and uncertainty-dependent evidence weighting. PerSeM relies on hand-designed trust and context cues; as indicated by the component analysis, reliability cues are not sufficient in isolation and require contextual constraints to avoid unstable corrections. Sequential trust replay introduces sensitivity to observation ordering and sampling density, motivating future order-robust temporal weighting. The framework further assumes accurate depth and camera poses, while our pose-noise analysis does not capture all effects of long-term localisation drift or correlated geometric errors.

Open-vocabulary mapping is further affected by mismatches between model predictions and closed-set benchmark taxonomies, as reflected by the gap between fine-grained and grouped UAVScenes evaluation. Future work will investigate learned uncertainty-aware refinement, adaptive history and context aggregation, explicit pose uncertainty, dynamic scenes, and scalable memory representations while preserving PerSeM's ability to selectively revisit ambiguous semantic states.

\section{Conclusion}

We presented PerSeM, a training-free persistent semantic memory framework for long-horizon open-vocabulary UAV mapping. PerSeM associates frame-wise semantic observations with persistent world-space voxels across repeated scene observations and conservatively refines majority memory through history-preserving aggregation, trust-aware evidence replay, and context-guided verification.

Experiments on Forest and UAVScenes show that persistent 3D aggregation provides the primary improvement over frame-wise predictions, while PerSeM provides consistent additional gains over the resulting strong memory baseline. Analysis of independently identified difficult regions further shows that these gains increase in semantically ambiguous and temporally unstable states, supporting selective refinement rather than global rewriting of persistent memory. These results suggest that a key challenge in long-horizon open-vocabulary mapping is not only aggregating repeated observations, but identifying when an established semantic memory should be reconsidered. PerSeM provides a lightweight mechanism for doing so without retraining or additional neural-network inference.

\bibliographystyle{unsrt}
\bibliography{references}

@String(CVPR= {IEEE Conf. Comput. Vis. Pattern Recog.})

@String(ICCV= {Int. Conf. Comput. Vis.})

@String(CVPR  = {CVPR})

@String(ICCV  = {ICCV})

@inproceedings{li2025segearthov,
  title={{SegEarth-OV: Towards Training-Free Open-Vocabulary Segmentation for Remote Sensing Images}},
  author={Li, Kaiyu and Liu, Ruixun and Cao, Xiangyong and Bai, Xueru and Zhou, Feng and Meng, Deyu and Wang, Zhi},
  booktitle={Proceedings of the IEEE/CVF Conference on Computer Vision and Pattern Recognition},
  pages={10545--10556},
  year={2025}
}

@inproceedings{kirillov2023segment,
  title={{Segment Anything}},
  author={Kirillov, Alexander and Mintun, Eric and Ravi, Nikhila and Mao, Hanzi and Rolland, Chloe and Gustafson, Laura and Xiao, Tete and others},
  booktitle={Proceedings of the IEEE/CVF International Conference on Computer Vision},
  pages={4015--4026},
  year={2023}
}

@inproceedings{bertasius2020maskprop,
  title={{Classifying, Segmenting, and Tracking Object Instances in Video with Mask Propagation}},
  author={Bertasius, Gedas and Torresani, Lorenzo},
  booktitle={Proceedings of the IEEE/CVF Conference on Computer Vision and Pattern Recognition},
  pages={9739--9748},
  year={2020}
}

@article{cheng2021rethinking,
  title={{Rethinking Space-Time Networks with Improved Memory Coverage for Efficient Video Object Segmentation}},
  author={Cheng, Ho Kei and Tai, Yu-Wing and Tang, Chi-Keung},
  journal={Advances in Neural Information Processing Systems},
  volume={34},
  pages={11781--11794},
  year={2021}
}

@inproceedings{li2022videoknet,
  title={{Video K-Net: A Simple, Strong, and Unified Baseline for Video Segmentation}},
  author={Li, Xiangtai and Zhang, Wenwei and Pang, Jiangmiao and Chen, Kai and Cheng, Guangliang and Tong, Yunhai and Loy, Chen Change},
  booktitle={Proceedings of the IEEE/CVF Conference on Computer Vision and Pattern Recognition},
  pages={18847--18857},
  year={2022}
}

@article{yue2024semanticnav,
  title={{Semantic-Driven Autonomous Visual Navigation for Unmanned Aerial Vehicles}},
  author={Yue, Pengyu and Xin, Jing and Zhang, Youmin and Lu, Yongchang and Shan, Mao},
  journal={IEEE Transactions on Industrial Electronics},
  volume={71},
  number={11},
  pages={14853--14863},
  year={2024}
}

@inproceedings{bartolomei2020perceptionaware,
  title={{Perception-Aware Path Planning for UAVs Using Semantic Segmentation}},
  author={Bartolomei, Luca and Teixeira, Lucas and Chli, Margarita},
  booktitle={2020 IEEE/RSJ International Conference on Intelligent Robots and Systems (IROS)},
  pages={5808--5815},
  year={2020},
  organization={IEEE}
}

@inproceedings{bartolomei2021semanticaware,
  title={{Semantic-Aware Active Perception for UAVs Using Deep Reinforcement Learning}},
  author={Bartolomei, Luca and Teixeira, Lucas and Chli, Margarita},
  booktitle={2021 IEEE/RSJ International Conference on Intelligent Robots and Systems (IROS)},
  pages={3101--3108},
  year={2021},
  organization={IEEE}
}

@article{xu2022visionaided,
  title={{Vision-Aided UAV Navigation and Dynamic Obstacle Avoidance Using Gradient-Based B-Spline Trajectory Optimization}},
  author={Xu, Zhefan and Xiu, Yumeng and Zhan, Xiaoyang and Chen, Baihan and Shimada, Kenji},
  journal={arXiv preprint arXiv:2209.07003},
  year={2022}
}

@inproceedings{mccormac2017semanticfusion,
  title={{SemanticFusion: Dense 3D Semantic Mapping with Convolutional Neural Networks}},
  author={McCormac, John and Handa, Ankur and Davison, Andrew and Leutenegger, Stefan},
  booktitle={2017 IEEE International Conference on Robotics and Automation (ICRA)},
  pages={4628--4635},
  year={2017},
  organization={IEEE}
}

@inproceedings{narita2019panopticfusion,
  title={{PanopticFusion: Online Volumetric Semantic Mapping at the Level of Stuff and Things}},
  author={Narita, Gaku and Seno, Takamasa and Ishikawa, Tomoya and Kaji, Yasuyuki},
  booktitle={2019 IEEE/RSJ International Conference on Intelligent Robots and Systems (IROS)},
  pages={4205--4212},
  year={2019},
  organization={IEEE}
}

@article{rosinol2021kimera,
  title={{Kimera: From SLAM to Spatial Perception with 3D Dynamic Scene Graphs}},
  author={Rosinol, Antoni and Abate, Marco and Chang, Yun and Carlone, Luca},
  journal={The International Journal of Robotics Research},
  volume={40},
  number={12-14},
  pages={1510--1546},
  year={2021}
}

@article{jatavallabhula2023conceptfusion,
  title={{ConceptFusion: Open-Set Multimodal 3D Mapping}},
  author={Jatavallabhula, Krishna Murthy and Kuwajerwala, Alihusein and Gu, Qiao and Omama, Mohd and Chen, Tao and Maalouf, Alaa and Li, Shuang and others},
  journal={arXiv preprint arXiv:2302.07241},
  year={2023}
}

@article{hughes2022hydra,
  title={{Hydra: A Real-Time Spatial Perception System for 3D Scene Graph Construction and Optimization}},
  author={Hughes, Nathan and Chang, Yun and Carlone, Luca},
  journal={arXiv preprint arXiv:2201.13360},
  year={2022}
}

@inproceedings{peng2023openscene,
  title={{OpenScene: 3D Scene Understanding with Open Vocabularies}},
  author={Peng, Songyou and Genova, Kyle and Jiang, Chiyu and Tagliasacchi, Andrea and Pollefeys, Marc and Funkhouser, Thomas},
  booktitle={Proceedings of the IEEE/CVF Conference on Computer Vision and Pattern Recognition},
  pages={815--824},
  year={2023}
}

@inproceedings{cheng2022xmem,
  title={{XMem: Long-Term Video Object Segmentation with an Atkinson-Shiffrin Memory Model}},
  author={Cheng, Ho Kei and Schwing, Alexander G.},
  booktitle={European Conference on Computer Vision},
  pages={640--658},
  year={2022},
  publisher={Springer Nature Switzerland}
}

@article{yang2021associating,
  title={{Associating Objects with Transformers for Video Object Segmentation}},
  author={Yang, Zongxin and Wei, Yunchao and Yang, Yi},
  journal={Advances in Neural Information Processing Systems},
  volume={34},
  pages={2491--2502},
  year={2021}
}

@inproceedings{huang2023vlmaps,
  title={{Visual Language Maps for Robot Navigation}},
  author={Huang, Chenguang and Mees, Oier and Zeng, Andy and Burgard, Wolfram},
  booktitle={2023 IEEE International Conference on Robotics and Automation (ICRA)},
  pages={10608--10615},
  year={2023},
  organization={IEEE}
}

@inproceedings{wang2025uavscenes,
  title={{UAVScenes: A Multi-Modal Dataset for UAVs}},
  author={Wang, Sijie and Li, Siqi and Zhang, Yawei and Yu, Shangshu and Yuan, Shenghai and She, Rui and Guo, Quanjiang and Zheng, JinXuan and Howe, Ong Kang and Chandra, Leonrich and others},
  booktitle={Proceedings of the IEEE/CVF International Conference on Computer Vision (ICCV)},
  pages={28946--28958},
  year={2025}
}

@article{blaga2024forestinspection,
  title={{Forest Inspection Dataset for Aerial Semantic Segmentation and Depth Estimation}},
  author={Blaga, Bianca-Cerasela-Zelia and Nedevschi, Sergiu},
  journal={arXiv preprint arXiv:2403.06621},
  year={2024}
}

@article{martins2025openvocabulary,
  author    = {Tomas Berriel Martins and Martin R. Oswald and Javier Civera},
  title     = {{Open-Vocabulary Online Semantic Mapping for SLAM}},
  journal   = {IEEE Robotics and Automation Letters},
  volume     = {10},
  number     = {11},
  pages      = {11745--11752},
  year       = {2025},
  doi        = {10.1109/LRA.2025.3617736}
}

@inproceedings{yamazaki2024openfusion,
  title={{Open-Fusion: Real-Time Open-Vocabulary 3D Mapping and Queryable Scene Representation}},
  author={Yamazaki, Kashu and Hanyu, Taisei and Vo, Khoa and Pham, Thang and Tran, Minh and Doretto, Gianfranco and Nguyen, Anh and Le, Ngan},
  booktitle={2024 IEEE International Conference on Robotics and Automation (ICRA)},
  pages={9411--9417},
  year={2024},
  publisher={IEEE},
  doi={10.1109/ICRA57147.2024.10610261}
}

@inproceedings{deng2025openvox,
  author    = {Yinan Deng and Bicheng Yao and Yihang Tang and Tianxing Zhou and Yi Yang and Yufeng Yue},
  title     = {{OpenVox: Real-time Instance-level Open-Vocabulary Probabilistic Voxel Representation}},
  booktitle = {Proceedings of the IEEE/RSJ International Conference on Intelligent Robots and Systems (IROS)},
  pages     = {1305--1311},
  year      = {2025},
  publisher = {IEEE}
}

@inproceedings{deng2026ovimap,
  author    = {Zilong Deng and Federico Tombari and Marc Pollefeys and Johanna Wald and Daniel Barath},
  title     = {{OVI-MAP: Open-Vocabulary Instance-Semantic Mapping}},
  booktitle = {Proceedings of the IEEE/CVF Conference on Computer Vision and Pattern Recognition (CVPR)},
  pages     = {12606--12616},
  year      = {2026}
}

@article{homberger2026fus3dmaps,
  author  = {Timon Homberger and Finn Lukas Busch and Jes{\'u}s Gerardo Ortega Peimbert and Quantao Yang and Olov Andersson},
  title   = {{FUS3DMaps: Scalable and Accurate Open-Vocabulary Semantic Mapping by 3D Fusion of Voxel- and Instance-Level Layers}},
  journal = {arXiv preprint arXiv:2605.03669},
  year    = {2026}
}

\end{document}